\documentclass{article}
\usepackage{spconf,amsmath,graphicx}
\usepackage{url}
\usepackage{hyperref}
\usepackage{tablefootnote}
\usepackage{threeparttable}
\usepackage{enumitem}
\usepackage{fancyhdr}

\newenvironment{myenum}{
\begin{enumerate}
 \setlength{\itemsep}{1pt}
 \setlength{\parskip}{0pt}
 \setlength{\parsep}{0pt}
}{\end{enumerate}}

\newenvironment{myitemize}{
\begin{itemize}
 \setlength{\itemsep}{1pt}
 \setlength{\parskip}{0pt}
 \setlength{\parsep}{0pt}
}{\end{itemize}}

\usepackage{xcolor}

\title{EXPLAINABLE FACT-CHECKING THROUGH QUESTION ANSWERING}
\name{Jing Yang$^1$, Didier Vega-Oliveros$^1$, Taís Seibt$^2$, and Anderson Rocha$^1$\ \thanks{Research funded by the São Paulo Research Foundation
(FAPESP) under the Grants DéjàVu \#2017/12646-3, \#2019/04053-8 and \#2019/26283-5.}}
\address{$^1$ Artificial Intelligence Lab. (\url{Recod.ai}), Institute of Computing, University of Campinas, SP, Brazil \\
$^2$ Department of Journalism, 
Universidade do Vale do Rio dos Sinos (Unisinos), 
Porto Alegre, RS, Brazil}

\fancypagestyle{firstpage}{
    \fancyhf{}
    \fancyfoot[L]{\footnotesize\textcopyright 2021 IEEE.  Personal use of this material is permitted. Permission from IEEE must be obtained for all other uses, in any current or future media, including reprinting/republishing this material for advertising or promotional purposes, creating new collective works, for resale or redistribution to servers or lists, or reuse of any copyrighted component of this work in other works.}
 }
 
\begin{document}
\ninept
\maketitle
\begin{abstract}

Misleading or false information has been creating chaos in some places around the world. To mitigate this issue, many researchers have proposed automated fact-checking methods to fight the spread of fake news. However, most methods cannot explain the reasoning behind their decisions, failing to build trust between machines and humans using such technology. Trust is essential for fact-checking to be applied in the real world. Here, we address fact-checking explainability through question answering. In particular, we propose generating questions and answers from \textbf{claims} and answering the same questions from \textbf{evidence}. We also propose an answer comparison model with an attention mechanism attached to each question. Leveraging question answering as a proxy, we break down automated fact-checking into several steps --- this separation aids models' explainability as it allows for more detailed analysis of their decision-making processes. Experimental results show that the proposed model can achieve state-of-the-art performance while providing reasonable explainable capabilities\footnote{Our implementation is available at: \url{https://bit.ly/3mtSUE3}}.

\end{abstract}
\section{Introduction}
\label{sec:intro}
The Internet has become an extension of our physical world, whereby almost everyone is connected. As a result, one small piece of false news can spread worldwide and ruin a person, company, or even a country's economy or reputation. To mitigate the impact of fake news, recognizing and flagging them is necessary due to the amount of news released. 

In the research community, many researchers have devoted efforts to studying automatic fake news detection. Among them, automated fact-checking has attracted a great deal of attention~\cite{thorne2018fever,liu2020fine}. The task is to check if a claim is factually correct based on evidence retrieved from reliable sources. However, according to a recent survey~\cite{nakov2021automated}, human fact-checkers generally do not trust automated solutions. Some works have been proposed to build the bridge between humans and machines. For example, Yang et al.~\cite{yang2021scalable} proposed a work to summarize claims for more scalable fact-checking and involved human-in-the-loop to evaluate summarization results.

Another popular line of work to increase trust for fact-checking is to generate explanations for the predicted results. Atanasova et al.~\cite{atanasova2020generating} first proposed a pioneer work to generate explanations. They performed an optimization to learn together veracity prediction and explanation extraction from evidence. Subsequently, Kotonya et al.~\cite{kotonya2020explainable} proposed a joint extractive and abstractive text summarization method for explanation generation. The authors also published a survey specifically about generating fact-checking explanations~\cite{kotonya2020survey}.

However, although generating explanations can provide more precise evidence to understand fact-checking decisions, existing systems lack a way to evaluate the explanations properly. Especially for explanations based on abstractive document summarization, researchers have shown that such models have problems of hallucination~\cite{cao2018faithful,maynez2020faithfulness}, generating summaries factually inconsistent with their original document. To deal with this issue, several works have been proposed~\cite{kryscinski2020evaluating, durmus2020feqa, wang2020asking}. In particular, Pagnoni et al. ~\cite{pagnoni2021understanding} summarized different types of errors some models make and metrics used to evaluate them. Among these evaluation metrics, leveraging question answering (QA) as a proxy has been the focus of some work~\cite{durmus2020feqa, wang2020asking}. The idea is to rely upon a question answering mechanism as an evaluation for the faithfulness of summaries.
Wang et al.~\cite{wang2020asking} extracted answers and questions from summaries and fine-tuned a QA model to generate answers from the documents; the answers from the document and its summary for the same questions are compared to determine the actual consistency of the summary. Recently, Nan et al.~\cite{nan2021improving} proposed an improved method than \cite{wang2020asking}, where instead of generating answers for both the summary and document, they model the likelihood of the summary and document conditioned on question-answer pairs generated from the summaries. Through this, the likelihood metric becomes suitable as a training objective to improve the factual consistency of summaries. 

A few works have been proposed to leverage QA to help in fact-checking. For example, in PathQG~\cite{wang2020pathqg}, Wang et al. generated questions from facts. They accomplished this task in two steps: first, they identified facts from an input text to build a knowledge graph (KG) and then generated an ordered sequence as a query path; second, they utilized a seq2seq model to learn to generate questions based on the query path. The human evaluation showed that their model could generate informative questions. In another work, Fan et al.~\cite{fan2020generating} generated question-answer pairs as a type of brief, along with passage brief and entity brief, and provided them to the human fact-checkers, aiming at improving their checking efficiency. 

Inspired by the QA works in checking factual consistency of documents and their summaries, we believe it is suitable for the fact-checking task, where we assess if claims are factually consistent with retrieved evidence. Therefore, we propose to leverage automated QA protocols and integrate them into the traditional fact-checking pipeline. As a result, we can provide explainable fact-checking results through question answering. The answer comparison model will predict a label and pinpoint the wrong part of a claim by showing which questions are more important for the decision. In this way, humans fact-checker can easily interpret the results and correct them if necessary. Our work differs from prior works~\cite{wang2020pathqg,fan2020generating} because we not only generate question-answer pairs but also fully integrate QA protocols in the fact-checking pipeline to automatically compare answers and predict their labels. We compare the proposed method with several baselines, achieving state-of-the-art results but with the critical feature of adding explainability to the fact-checking process.

We summarize our contributions as follows.
\begin{myitemize}
    \item We propose a novel pipeline for using question answering as a proxy for explainable fact-checking;
    \item We introduce an answer comparison model with an attention mechanism on questions to learn their importance on the claims;
\end{myitemize}


\section{Proposed Methodology}
\label{sec:method}
We introduce question answering (QA) in the fact-checking process. Despite previous mentions of using QA for fact-checking, no previous work has explored integrating QA protocols in its pipeline. Our proposed solution is described as follows:
\begin{myenum}
    \item[1)] Given a claim $C$, generate multiple questions ${Q_1}$,$\cdots$, ${Q_n}$ and answers $A^C_1$, $\cdots$, $A^C_n$ from it;
    \item[2)] Retrieve and re-rank evidence $E$ based on the claim (and possibly questions); 
    \item[3)] For each question generated from 1), ask retrieved evidence for answers $A^E_1$, $\cdots$, $A^E_n$ respectively;
    \item[4)] Compare the answer pairs $(A^C_i, A^E_i)$ and transform the result into a label of SUPPORTS or REFUTES.
\end{myenum}

Our proposed pipeline leads to more explainability as we break down the fact-checking process into more steps, allowing a more fine-grained analysis of each part of the process (e.g., question generation, question answering, or answer comparison). In addition, through answer generation from claims and evidence, we vastly reduce the information (from claims and their evidence to only answer pairs) fed to the final classification model. Thus, the model learns from more direct and precise inputs. 

To focus on how question answering empowers explainability, we use gold evidence instead of retrieved evidence. It means that for step 2), we take the gold evidence directly instead of retrieving them to focus on evaluating the other three stages of the problem. Future work will be dedicated to the retrieval by itself. Therefore, we focus on steps 1), 3), and 4) of the pipeline. Next, we detail the proposed methodology steps.

\subsection{Question and Answer Generation}
To generate questions from a text, answers for the text are usually provided first to generate more relevant questions~\cite{durmus2020feqa,wang2020asking}. Answers are usually extracted based on named entities and noun phrases; then, questions are generated given the claim and answers. They can also be generated in parallel with questions~\cite{nan2021improving}. We adopt the approach to generate questions and answers from claims simultaneously~\cite{nan2021improving}. In particular, we follow the instruction of~\cite{nan2021improving} to fine-tune the BART-large model to generation question-answer pairs $(Q_1, A^C_1)$, $\cdots$, $(Q_n, A^C_n)$ from a given claim $C$.  Using beam search, 64 question-answer pairs are generated, then pairs are removed if the claim does not contain the answers.
For answers of evidence $E$, we utilize a pre-trained extractive QA model to answer the questions generated previously from the claim. The model generates multiple answers, and we choose the one with the highest score (the most likely answer). 

\subsection{Answer Pair Comparison}

\begin{figure}[t]
        \centering
        \includegraphics[width=0.49\textwidth]{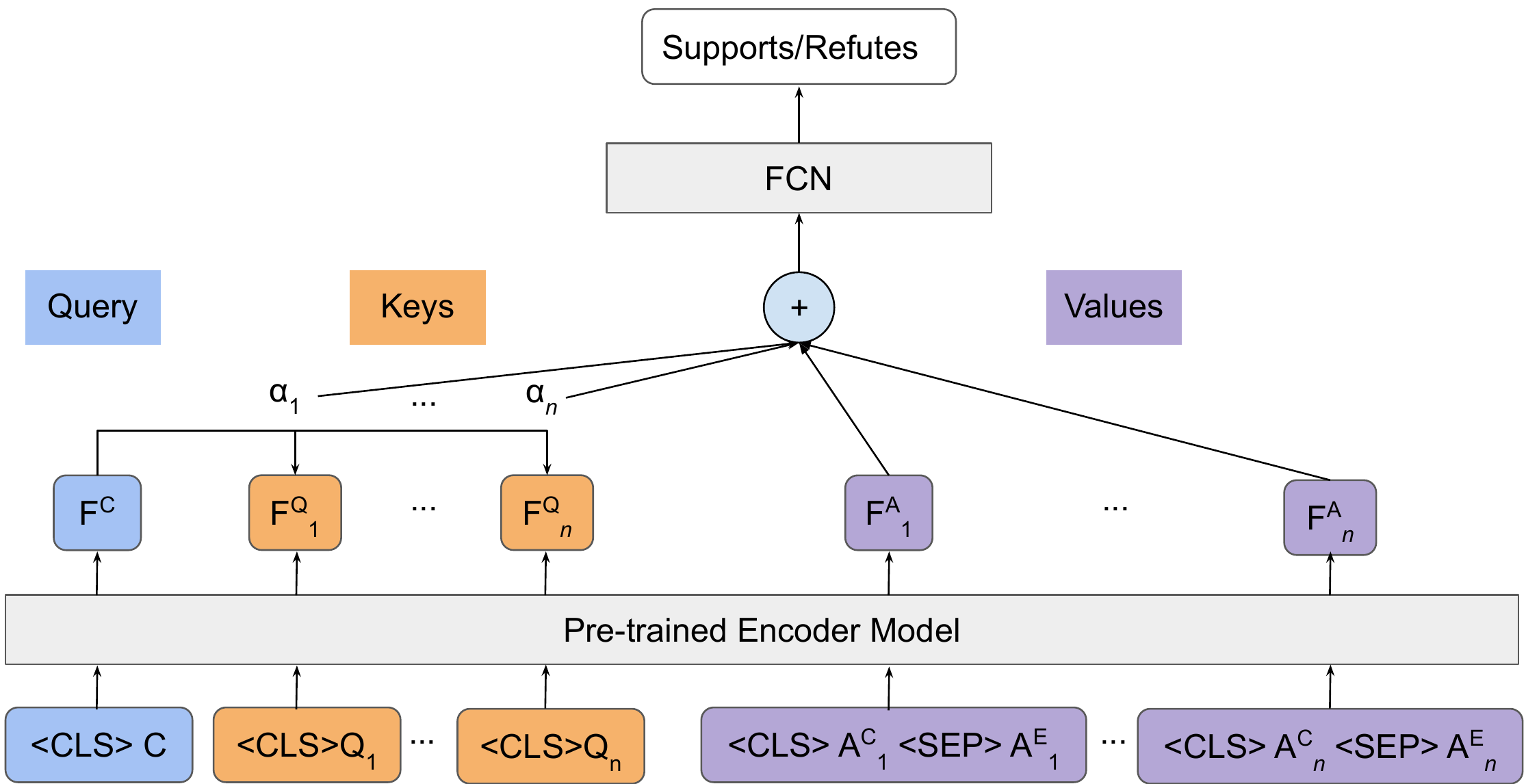}
        \caption{Answer comparison model with attention on questions. $C$ represents a given claim, $Q_i$ represents $i_{th}$ questions, and $({A^C}_i, {A^E}_i)$ represents $i_{th}$ answer pairs for claim and evidence. $n$ denotes the number of questions and answer pairs.}
        \label{f:attention}
    \end{figure}

For answer comparison, the token-level F1 score is usually used to measure similarity between answer pairs; however, it does not work when the two answers have non-overlapping words but are semantically similar. We propose to fine-tune a transformer model to learn answer comparison. Considering that different questions have various purposes, thus also vary in their importance. To account for this, we add attention to each question to learn the importance weights. The structure of the model is shown in Fig.~\ref{f:attention}.

Specifically, we rely on a pre-trained masked language model to encode the claim $C$, questions $Q_1$, $\cdots$, $Q_n$, and answer pairs $({A^C}_1, {A^E}_1)$, $\cdots$, $({A^C}_n, {A^E}_n)$. For the answer pairs, we add a \textlangle SEP\textrangle ~token between two answers of the same question. We use one encoder model for encoding all the inputs, which means the weights are shared. After the encoding, we take the representation of the \textlangle CLS\textrangle ~token as each sentence embedding, thus transforming the claim, questions and answer pairs into features: $F_C$, $F^Q_1$, $F^Q_2$, $\cdots$, $F^Q_n$, and ${F^A_1}$, $F^A_2$, $\cdots$, $F^A_n$ respectively, where $n$ is the number of questions for each claim. Then we utilize additive attention proposed by Bahdanau et al.~\cite{bahdanau2015neural} to learn the importance of each question. We treat the claim as a query, questions as keys, and answers as values for each representation. The details are formulated as follows. 
\begin{align}
\begin{split}\label{eq:1}
f_\text{att}\left({F^C}, {F^Q}_{j}\right)&=\mathbf{W}_{3} \tanh \left(\mathbf{W}_{1} {F^C}+\mathbf{W}_{2} {F^Q}_{j}\right)
\end{split}\\
\begin{split}\label{eq:2}
a_j &= \text{softmax}(f_\text{att}\left({F^C}, {F^Q}_{j}\right))
\end{split}\\
\mathbf{F}&=\sum_{j} a_{j} F^A_{j}  \label{eq:3}
\end{align}
 where $f_{a t t}$ calculate the attention weight between ${F^C}$ and $F^Q_j$ ($j = 1, 2, \cdots, n$), $\mathbf{W}_{1}$, $\mathbf{W}_{2}$ and $\mathbf{W}_{3}$ are learnable parameters. 
 
In Eq. (\ref{eq:1}) and (\ref{eq:2}), attention weights are calculated and normalized by the softmax function. Then Eq. (\ref{eq:3}) uses a weighted sum to combine all answer features to the final feature $F$. Feature $F$ is then fed to a fully connected layer to have the final prediction of SUPPORTS or REFUTES. Notice that all the information present in both claim and evidence are reduced into their respective answers. The claim and generated questions take part in selecting the most relevant answers, assigning higher weights to them.

\section{Experimental Setup}
\label{sec:setup}
 
\subsection{Dataset}
 We adopt the Fool-Me-Twice (FM2) dataset, which comprises 12,968 claims and their associated evidence. FM2 is a recently published dataset collected through a multi-player game. In the game, one player generates a claim and tries to fool other players. The others have to decide if the claim is true or false based on evidence retrieved by the game before a timer runs out. The game's setting makes this dataset challenging as the players are motivated to generate claims hard to verify. 
 FM2 a more difficult and less biased dataset than the seminal dataset FEVER~\cite{thorne2018fever}, in which a model can exploit specific words from the claim~\cite{schuster2019towards} to achieve reasonable accuracy (79.1\% for two classes). In contrast, FM2 is shown not to have biases, a classification based only on claims resulted in low prediction accuracy (61.9\%).

\subsection{Implementation details}
For question-answer pairs generation, we follow the code provided in ~\cite{fan2020generating} to fine-tune a BART-large model based on XSUM and CNNDM datasets\footnote{{\url{https://bit.ly/3iBZyqR}}}. For answer generation for evidence, we use the FARM framework from deepset\footnote{\url{https://github.com/deepset-ai/FARM}} to generate answers from evidence, and the question-answering model is \textit{deepset/electra-base-squad2}. For answer comparison, we use \textit{microsoft/mpnet-base} model for encoding all input representation, as it has shown to perform well in question answering tasks~\cite{song2020mpnet}. As the question generation model does not output the same number of questions for every claim, we selected the first ten questions if the claim has more than 10; if the number of questions is less than 10, we repeat the first question until 10. We choose this quantity of questions because the average number of questions for each claim is 11.5. The hyperparameters for training the answer comparison models are: \textit{number of epochs} = 5,  \textit{batch size} =32, \textit{learning rate} = 2e-5, which is the standard for fine-tuning a masked language model, and \textit{maximum token length} = 32. For statistical significance, we run each experiment 5 times and report the average and standard deviation. As the dataset is well-balanced, we use macro average accuracy as the evaluation metric.

\subsection{Baselines}
We set questions and answers for the baselines to be the same, only varying different answer comparison methods. 

\begin{itemize}[leftmargin=*]
    \item \textbf{Blackbox method:} we compare our results with the original proposed method in \cite{eisenschlos2021fool}. We refer to it as the \textit{black box method} as they concatenate claim and evidence for the prediction without providing interpretability. We used the code provided by the authors\footnote{\url{https://bit.ly/2ZO6CtR}} and ran it five times to have an average result.
    \item \textbf{QUALS score:} it is an automatic metric for checking factual consistency~\cite{nan2021improving}. It does not generate answers for evidence. Instead, it calculates the likelihood of the evidence given the question-answer pair from the claim, compromising explainability.  
    \item \textbf{Token level F1-score:} a standard metric for question-answer tasks. It counts words overlap between two answers.
    \item \textbf{BERTscore:} a common metric for measuring the similarity of two sentences. We use the default model~\textit{roberta-large}.
    \item \textbf{Cosine similarity:} a metric also used for measuring sentence similarity. We use sentence transformer \textit{all-mpnet-base-v2} to embed the answers and calculate the cosine similarities between the embeddings. 
\end{itemize}

Only the black box method requires training. 
The others are metrics to evaluate the answer pairs. These metrics calculate a score representing similarity for each answer pair, except for QUALS that outputs a score for all answers of the same claim.
As each claim has several questions, we compute the average score for the claim and provide a threshold to convert the score to a binary label.

\section{Results and Analysis}
\label{sec:result}

\subsection{Comparison with baselines}
We show the results with different baselines in Table~\ref{t:result}. For the metric-based methods, we do a binary search to find the highest accuracy on the development set for the threshold selection. 

\begin{table}[t]
\centering
\caption{Fact-checking label accuracy of different methods. `X-AI' denotes Explainability capabilities.}
\footnotesize
\begin{tabular}{lrr}
\hline
Methods               & Dev Acc & Test Acc \\ \hline
Blackbox (No X-AI)            & \textbf{76.17$\pm$1.23}   & \textbf{74.58$\pm$1.66} \\
QUALS (th=-1.2)       & 56.12   & 56.01    \\
BERTscore (th=0.843)  & 58.68   & 62.32    \\
cosine sim (th=0.305) & 61.16   & 62.75    \\
F1-score (th=0.06)    & 64.07   & 63.77    \\
\hline
\textcolor{black}{Attention C-Q-AA (ours, X-AI)}     & \textcolor{black}{\underline{75.44$\pm$0.52}}   & \textcolor{black}{\underline{73.43$\pm$0.83}}  \\ \hline
\end{tabular}
    \begin{tablenotes}
      \item th: threshold
    \end{tablenotes}
\label{t:result}
\end{table}

The results show that training an answer comparison model specifically for the fact-checking task improves accuracy compared with the methods without training. Our attention-based method achieves slightly lower accuracy than the black box method. However, our method is more suitable for real-world applications than the black box one because: 1) our method essentially reduces the input needed for prediction while remaining almost the same accuracy, 2) we enable error analysis for fact-checking with several steps, and 3) our model additionally provides more explainability by learning the importance of each question.

\subsection{Attention visualization}
\begin{figure}[t]
        \centering
        \includegraphics[width=0.45\textwidth, height=0.4\textwidth]{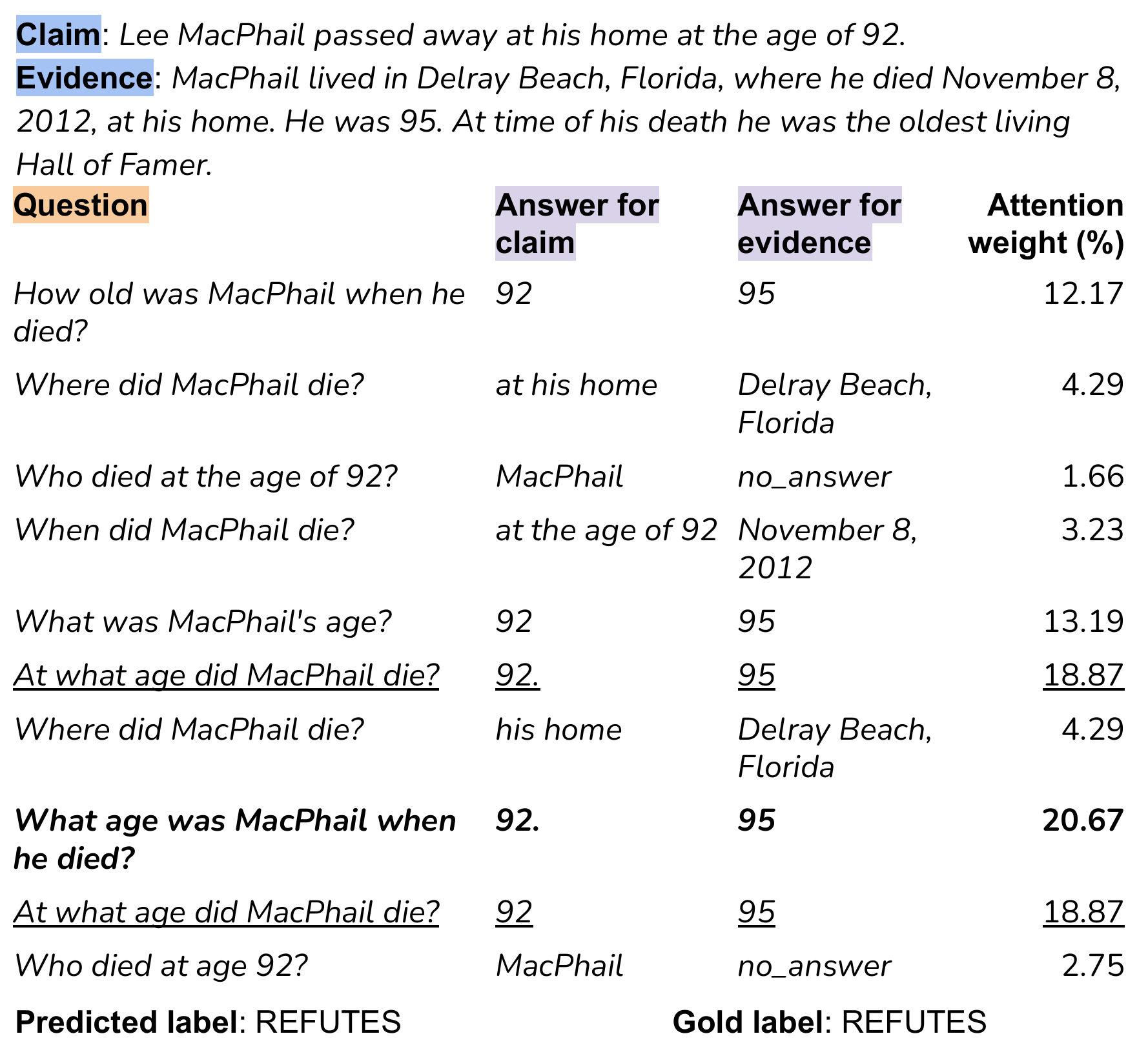}
        \caption{An example of our model generated questions, answer pairs, and attention weights. The question with the highest weight is in bold, and the second highest is underlined. }
        \label{f:refute}
    \end{figure}

    To illustrate how attention helps explainability, we show an example of our generated questions with attention weights and their answers from claim and evidence in Fig.~\ref{f:refute}.
    The question with the highest weight is bold, and the second-highest underlined. Although some answers are incorrect and there are non-matching answer pairs, the model can attend more on the questions and answers more relevant to the factuality of the claim, showing our approach's potential. We also see that because the claim is short, most questions are repetitive.

\subsection{Ablation study}
\begin{figure}[h]
        \centering
        \includegraphics[width=0.3\textwidth]{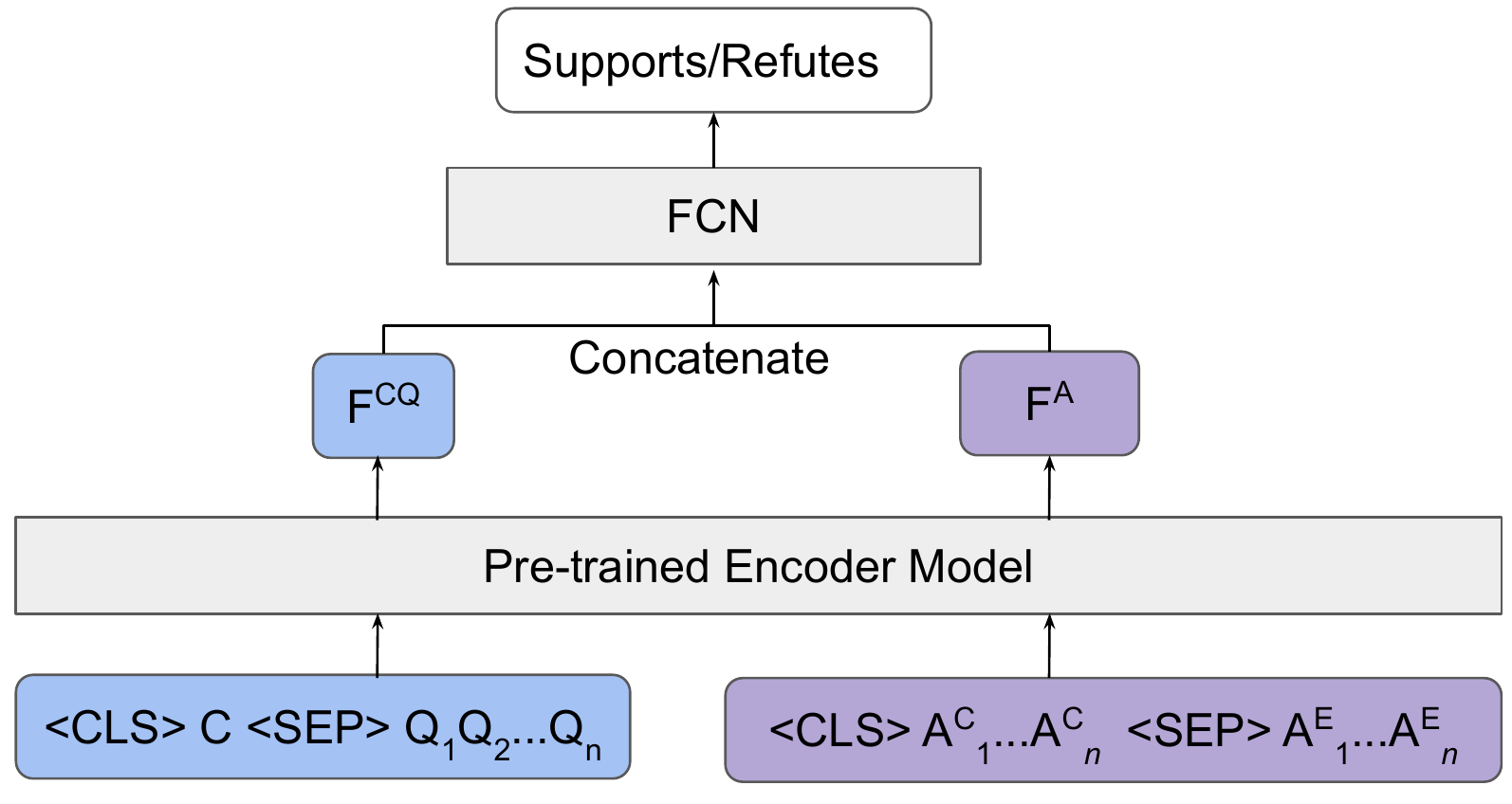}
        \caption{Answer comparison model without attention on questions. $C$ represents a given claim, $Q_i$ represents $i_{th}$ questions, and $({A^C}_i, {A^E}_i)$ represents $i_{th}$ answer pairs for claim and evidence. $n$ denotes the number of questions for claim $C$.}
        \label{f:noatte}
    \end{figure}

We carry out an ablation study to show if our attention mechanism improves performance compared with simple classification. Thus we remove the attention layer of our proposed attention model, the network structure is shown in Fig.~\ref{f:noatte}. Specifically, to use all available questions, we concatenate all questions and all answers: so the model has two inputs \textlangle CLS\textrangle~$C$ \textlangle SEP\textrangle~$Q_1Q_2$ $\cdots$ $Q_n$, and \textlangle CLS\textrangle~$A^C_1 A^C_2$  $\cdots$ $A^C_n$  \textlangle SEP\textrangle~$A^E_1 A^E_2$ $\cdots$ $A^E_n$ (note here $n$ can be different for different claims). As the inputs are concatenated, the maximum token length here is 128. Then through the embedding model, each input is transformed into a feature vector, and the two vectors are concatenated to be fed into the classification layer. 

To study the effect of different components of our proposed model, we design the inputs as follows: 

\begin{myitemize}
    \item C: only claims \textlangle CLS\textrangle~$C$;
    \item Q: only concatenated questions \textlangle CLS\textrangle~$Q_1Q_2 \cdots Q_n$;
    \item AA: only answer pairs \textlangle CLS\textrangle~$A^C_1 A^C_2$  $\cdots$ $A^C_n$\textlangle SEP\textrangle~$ A^E_1 A^E_2$ ~$\cdots$$A^E_n$;
    \item Q-AA:  concatenated questions \textlangle CLS\textrangle~$Q_1Q_2 \cdots Q_n$ and answer pairs \textlangle CLS\textrangle~$A^C_1 A^C_2  \cdots A^C_n$\textlangle SEP\textrangle~$ A^E_1 A^E_2 \cdots A^E_n$;
    \item CQ-AA: our full model without attention (shown in Fig.~\ref{f:noatte}).
    \item Attention C-Q-AA: our full model with attention.
\end{myitemize}
    
\begin{table}[t]
\centering
\caption{Ablation study of the model without attention}
\resizebox{0.35\textwidth}{!}{
\begin{tabular}{lrr}
\hline
Inputs      & Dev Acc & Test Acc \\ \hline
C     & 59.15$\pm$1.22   & 61.57$\pm$1.67    \\
Q     & 56.22$\pm$1.37   & 56.90$\pm$0.90    \\
AA    & 74.15$\pm$1.33   & 72.61$\pm$1.04    \\
Q-AA  & 74.46$\pm$0.80   & 72.62$\pm$1.59    \\
CQ-AA & 74.88$\pm$0.81   & 72.89$\pm$1.14    \\ 
Attention C-Q-AA & \textbf{75.44$\pm$0.52}    &   \textbf{73.43$\pm$0.83}   \\ \hline
\end{tabular}
}
\label{t:ablation}
\end{table}

From the ablation study, we want to know how much each input affects the model's performance. In Table~\ref{t:ablation}, we can see that with C or Q only, the model cannot perform well, indicating that the model can not rely solely on the claim information to achieve high accuracy. Our result agrees with the original paper, in which the model only with claims achieved an accuracy of 61.9\%. Also, when adding Q and CQ information to AA, Q-AA and CQ-AA perform slightly better. This indicates that the model can learn most of the information from answer pairs only. It is reasonable because the answer pairs carry most of the critical information from both claims and evidence. Comparing CQ-AA with our attention-based C-Q-AA, we see that the attention mechanism can help increase performance because it uses claims and questions to weigh up essential answer pairs.

\subsection{Limitations}    

\begin{itemize}[leftmargin=*]
    \item \textbf{Question Generation.} Generating diverse and relevant questions aiming at the factuality of a claim is challenging. Claims can be altered by changing the subject, object, time, place, actions, or even multiple editions together. In some cases, we observed that the questions have the problem of not recognizing complete phrases of the claim, and sometimes most questions of a claim are semantically similar because the claim is too short. For example in Fig.~\ref{f:refute}, we can see that most of the questions are paraphrases. Hence, better ways of generating questions and filtering less relevant and repetitive questions are needed to improve performance. 

    \item \textbf{Question Answering.} Answering correctly giving the context is a non-trivial and crucial step in the pipeline. Unfortunately, state-of-the-art models can fail to answer correctly in some cases, as they require reasoning and logical thinking to calculate the correct answer from the context. We show a failing example here: 
    \textbf{Evidence}: \textit{Weber was born in Eutin, Bishopric of Lübeck, the eldest of the three children of Franz Anton von Weber and his second wife, Genovefa Weber, a Viennese singer.} \textbf{Question}: \textit{How many siblings did Albert Weber have?} \textbf{Answer for evidence}: \textit{three}. 
    
    In the example, the model is not able to give the correct answer --\textit{two}, because it is an extractive QA model, which is a limitation of this type of model. Nevertheless, the explainability provide by questions and answers gives us a better idea of which part is wrong in the claim and what could help us improve the model. 
\end{itemize}
    
Reasoning over text is a very challenging task; other ways of transforming the claim into a format like tabular data~\cite{gupta2021my} may also help simplify the reasoning and thus improve performance.
    
\section{Conclusion}
\label{sec:conclusion}

In this paper, we proposed a novel pipeline for using QA as a proxy for fact-checking. Based on this pipeline, we proposed an answer comparison model with an attached attention mechanism, which learns to attend critical questions with interpretability capabilities. 

Our ablation study showed that the model can achieve near state-of-the-art performance with only information from answer pairs. Thus, using QA, we can encourage the model to learn from more precise evidence; this can aid fact-checkers in better understanding models' decisions. Then, when necessary, they can compare the answers and make decisions for themselves.

In future work, we plan to add the retrieval step to the pipeline instead of using gold evidence, as the retrieval is also a crucial part of fact-checking. We can also instead answer questions directly from a more extensive set of document evidence. In addition, we plan to work on more datasets to address the generalization capabilities of the method. Finally, we intend to have human evaluations on the questions and answers to improve the generation and potentially help human fact-checkers by providing high-quality QAs.

\bibliographystyle{IEEEbib}
\bibliography{refs}

\end{document}